\documentclass[10pt,letterpaper]{article}

\usepackage{cogsci_template}
\usepackage{cogsci}
\cogscifinalcopy
\hypersetup{
    citecolor=gray
}
\usepackage{CJKutf8}
\usepackage{multirow}
\usepackage{array}

\acrodef{bo}[BO]{Base Object}
\acrodef{iei}[IEI]{Identification-Explanation-Implication}
\acrodef{ccp}[CCP]{Combinational Creativity Product}
\acrodef{ai}[AI]{Artificial Intelligence}
\acrodef{llm}[LLM]{Large Language Model}
\acrodef{mllm}[MLLM]{Multimodal Large Language Model}
\acrodef{vlm}[VLM]{Vision-Language Model}
\newcommand{\dataset}{CreativeMashup}

\newcommand{\baseline}{\textbf{II}}
\newcommand{\cc}{\textbf{IEI}}

\newcommand{\one}{\ding{202}\xspace}
\newcommand{\two}{\ding{203}\xspace}
\newcommand{\three}{\ding{204}\xspace}

\title{Probing and Inducing Combinational Creativity in Vision-Language Models}
 
\author{%
    Yongqian Peng$^{1,2*}$, Yuxi Ma$^{1*}$, Mengmeng Wang$^{3}$, Yuxuan Wang$^{3}$,
    \vspace{3pt}\\
    \textbf{Yizhou Wang$^{4}$, Chi Zhang$^{1}$, Yixin Zhu$^{1,~\textrm{\Letter}}$, Zilong Zheng$^{3,~\textrm{\Letter}}$}
    \vspace{3pt}\\
    \small $^1$ Institute for Artificial Intelligence, Peking University\quad{}
    \small $^2$ Yuanpei College, Peking University\\
    \small $^3$ State Key Laboratory of General Artificial Intelligence, BIGAI\\
    \small $^4$ Center on Frontiers of Computing Studies, School of Computer Science, Peking University\\
    \small $^\star{}$Equal contributors \quad $\textrm{\Letter}$\,\,\texttt{yixin.zhu@pku.edu.cn,\,zlzheng@bigai.ai}
}

\begin{document}
\maketitle

\begin{abstract}
The ability to combine existing concepts into novel ideas stands as a fundamental hallmark of human intelligence. Recent advances in \acp{vlm} like GPT-4V and DALLE-3 have sparked debate about whether their outputs reflect combinational creativity---defined by \citet{boden1998creativity} as synthesizing novel ideas through combining existing concepts---or sophisticated pattern matching of training data. Drawing inspiration from cognitive science, we investigate the combinational creativity of \acp{vlm} from the lens of concept blending. We propose the \textbf{\ac{iei} framework}, which decomposes creative processes into three levels: identifying input spaces, extracting shared attributes, and deriving novel semantic implications. To validate this framework, we curate \dataset{}, a high-quality \textbf{dataset} of 666 artist-generated visual mashups annotated according to the \ac{iei} framework. Through extensive experiments, we demonstrate that in \textbf{comprehension} tasks, best \acp{vlm} have surpassed average human performance while falling short of expert-level understanding; in \textbf{generation} tasks, incorporating our \ac{iei} framework into the generation pipeline significantly enhances the creative quality of \acp{vlm}' outputs. Our findings establish both a theoretical foundation for evaluating artificial creativity and practical guidelines for improving creative generation in \acp{vlm}. Project page: \url{https://ppyyqq.github.io/aicc/}.

\textbf{Keywords:} creativity; combinational creativity; \acp{vlm}
\end{abstract}

\section{Introduction}

\begin{quote}
    \textit{Creativity is just connecting things.}
    
    \vspace{-2pt}\raggedleft --- Steve Jobs
\end{quote}\vspace{-2pt}

\begin{figure}[t!]
    \centering
    \includegraphics[width=\linewidth]{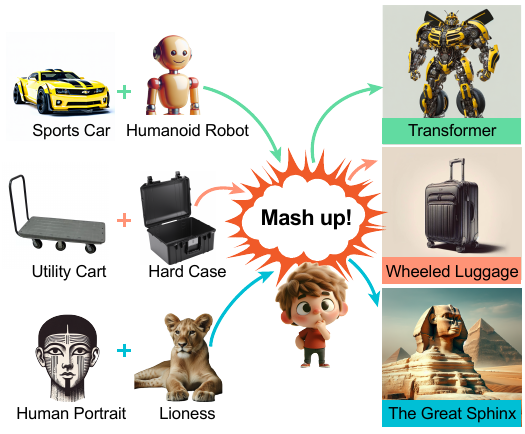}
    \caption{\textbf{Combinational creativity across domains.} Examples showing how combining two distinct elements creates novel concepts: sports car $+$ humanoid robot $\rightarrow$ Transformer (entertainment), utility cart $+$ hard case $\rightarrow$ wheeled luggage (industrial design), and human portrait $+$ lioness $\rightarrow$ Great Sphinx (ancient architecture). Each combination demonstrates how merging basic elements generates innovative outcomes.}
    \label{fig:teaser}
\end{figure}

Creativity, a defining characteristic of human intelligence, enables the production of novel concepts, solutions, and artistic expressions \citep{mehrotra2024enhancing,boden1998creativity,holyoak1996mental}. At its core, combinational creativity---the ability to generate new ideas by meaningfully combining familiar ones---represents one of the most fundamental creative processes \citep{han2019three,boden2009creativity}, influencing domains from art and design to scientific discovery \citep{guzdial2018combinatorial}; see examples in \cref{fig:teaser}. This uniquely human capacity allows us to actively shape our environment rather than merely respond to it \citep{gabora2010evolutionary}.

Recent advances in \acp{vlm} have demonstrated increasingly sophisticated generative capabilities across multiple modalities \citep{franceschelli2024creativity,chakrabarty2024art,bubeck2023sparks,ma2023brain}, producing creative content that closely resembles human-generated work \citep{orwig2024language,koivisto2023best,tian2024large,mehrotra2024enhancing}. However, this apparent success raises fundamental questions about the underlying mechanisms: while existing research has examined \acp{vlm}' creativity through specific lenses such as semantic association \citep{chen2023probing} and divergent thinking \citep{bellemare2024divergent}, we lack a systematic framework for evaluating whether these models implement genuine combinational creative processes or merely leverage statistical patterns in their training data.

To address this gap, we investigate \acp{vlm} through the theoretical framework of combinational creativity. Grounded in cognitive science, this framework provides explicit criteria for evaluating both creative processes and their outcomes. Specifically, we address two complementary research questions:
\one How effectively can \acp{vlm} \textit{understand} and \textit{interpret} combinational creative processes and products? 
\two Can the explicit incorporation of the combinational creative thinking process enhance the \textit{generative} capabilities of these systems?

\begin{figure*}[t!]
    \centering
    \includegraphics[width=\linewidth]{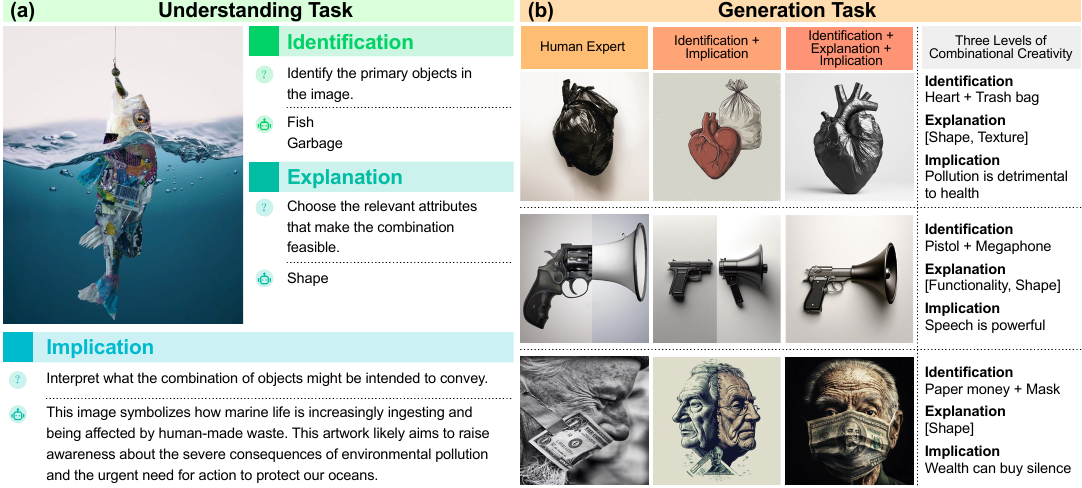}
    \caption{\textbf{Examples of the comprehension task and generation task.} (a) The understanding task demonstrates three evaluation components using a fish-garbage mashup image: human participants or \acsp{vlm} identify primary objects, explain combination attributes, and interpret implications. (b) The generation task compares outputs from human experts and two model settings (Identification + Implication \vs Identification + Explanation + Implication) across three concept pairs (heart-trash, pistol-megaphone, paper money-mask).}
    \label{fig:task} 
\end{figure*}

Drawing inspiration from conceptual blending theory \citep{fauconnier2003conceptual}, we introduce the \acf{iei} framework---a hierarchical decomposition of combinational creativity into three levels:
\begin{itemize}[leftmargin=*,noitemsep,nolistsep]
    \item \textbf{Identification}: Recognition of constituent elements (\textit{What objects are combined?})
    \item \textbf{Explanation}: Analysis of combinatorial mechanisms (\textit{How are these objects combined?})
    \item \textbf{Implication}: Interpretation of semantic meaning (\textit{What message does this combination convey?})
\end{itemize}
For comprehension tasks (Q\one), this framework enables systematic analysis through progressively deeper analytical levels. For generation tasks (Q\two), we investigate two approaches while maintaining consistent input specifications: a method that explicitly incorporates conceptual attribute mapping based on our \ac{iei} framework, and a baseline approach using standard chain-of-thought prompting \citep{wei2022chain}.

To validate this framework, we curate \dataset{}, a novel benchmark dataset designed for combinational creativity. \dataset{} features professional artists' original visual mashups paired with expert annotations that decompose each creation according to the \ac{iei} framework's three levels, providing ground-truth data for evaluating both comprehension and generation capabilities. Through extensive experiments on \dataset{}, we demonstrate that state-of-the-art \acp{vlm} achieve above-average human performance in \textbf{comprehension} tasks while lagging behind the expert performance, and that the explicit incorporation of our framework significantly enhances their creative \textbf{generation} capabilities.

\section{Related Work}

\paragraph{Combinational Creativity} 

Margaret A. Boden's seminal work \citep{boden1998creativity,boden2004creative,boden2009creativity} identifies three fundamental types of creativity, with combinational creativity---the generation of novel ideas through familiar combinations---being particularly relevant to \ac{ai} research. For \ac{ai} systems to achieve this form of creativity, Boden argues they must develop three critical capabilities: rich associative memory, understanding of human values, and computational expression of these values \citep{peng2024tong}. Despite recent advances in \ac{ai}, these fundamental aspects remain largely unexplored in modern research \citep{boden1998creativity,boden2004creative,ma2023brain}.

\paragraph{Conceptual Blending Theory}

Conceptual blending theory, developed in cognitive science, provides a theoretical framework for understanding combinational creativity \citep{fauconnier2003conceptual}. In its basic form, the theory describes four interconnected spaces: two input concept spaces, a generic space containing shared structures, and a blend space where new meanings emerge. The generic space serves as a crucial bridge, identifying common elements between input spaces that enable meaningful integration and the creation of novel insights \citep{fauconnier2003conceptual}. Early computational implementations, such as \citet{han2018combinator}'s system, approached this challenge via hand-authored semantic networks, combining related concepts in structured ways. 

\paragraph{Creativity in \acp{vlm}}

The emergence of pre-trained \acp{vlm} has demonstrated remarkable creative capabilities, raising fundamental questions about their relationship to human cognition and creative processes. Recent ``probing'' studies have employed both classical psychological measures and specialized methodologies to investigate \acp{vlm}'s creativity \citep{franceschelli2024creativity,tian2024large,koivisto2023best,hessel2023androids}. For instance, \citet{koivisto2023best}'s evaluated \ac{ai} chatbots using the Alternative Uses Task and revealed that while \ac{ai} systems generally outperformed average humans, the highest-quality human responses remained superior. Similarly, specialized evaluations (\eg, \citet{akula2023metaclue}'s visual metaphor dataset and \citet{ji2022abstract}'s Tangram puzzle assessments) have provided insights into specific aspects of machine creativity and abstract reasoning.

However, existing research primarily focuses on performance metrics rather than \textit{underlying creative mechanisms}. The proposed \ac{iei} framework addresses this limitation by providing a systematic approach to evaluating both the \textit{comprehension} and \textit{generation} aspects of combinational creativity in \acp{vlm}, moving beyond simple performance measures.

\section{The \texorpdfstring{\ac{iei}}{} Framework}

Drawing from conceptual blending theory \citep{fauconnier2003conceptual}, we decompose combinational creativity into three hierarchical levels, each corresponding to progressively deeper analytical capabilities.

\paragraph{Identification}

This foundational level involves recognizing and isolating constituent \textit{elements} that form potential combinations. Corresponding to input mental spaces in conceptual blending theory, this process requires the system not only to detect visual elements but also to understand their categorical identity and potential roles in combinations.

\paragraph{Explanation}

This intermediate level analyzes structural and semantic alignments between identified elements, mapping to the generic space in blending theory. This process encompasses (i) attribute extraction, identifying salient features of each element (\eg, shape, functionality); (ii) cross-space mapping, finding corresponding features between elements; and (iii) compatibility assessment, evaluating which shared attributes could support meaningful combinations.

\paragraph{Implication}

The most sophisticated level involves deriving emergent meaning from the combinations, corresponding to the blend space in conceptual blending theory. It requires (i) pattern completion, inferring implied relationships beyond explicit combinations; (ii) semantic integration, synthesizing coherent meaning from combined elements; and (iii) cultural/contextual reasoning, understanding broader implications within social/cultural contexts.

The \ac{iei} framework systematically evaluates combinational creativity in comprehension and generation tasks. For comprehension, each level within the \ac{iei} framework represents progressively complex analytical capabilities, offering clear metrics to investigate comprehension capabilities. In generation tasks, the \ac{iei} framework guides the creative process systematically, facilitating more sophisticated and meaningful outputs compared to standard chain-of-thought approaches. This structured decomposition supports the design of experiments that explicitly integrate systematic creative thinking, enhancing \acp{vlm}' capacity for contextually relevant creation.

\section{Experiment 1: Do \texorpdfstring{\acp{vlm}}{} Understand Combinational Creativity?}

We design three comprehension tasks (see \cref{fig:task}\textcolor{red}{a}) to investigate: To what extent can \acp{vlm} understand combinational creativity expressed in images?

\subsection{\dataset{} Dataset Construction}

To enable rigorous evaluation of combinational creativity comprehension, we developed \dataset{}, a carefully curated benchmark dataset. Working with professional artists, we collected 666 visual mashups specifically designed to exemplify creative combinations. Each mashup features the deliberate blending of two ordinary physical objects against simple backgrounds, allowing a clear focus on the creative combinations.

The distinguishing feature of the dataset is its comprehensive annotation scheme aligned with our \ac{iei} framework. Expert annotators provided detailed labels across all three levels: (i) identification-level annotations of constituent objects, (ii) explanation-level documentation of combining attributes, and (iii) implication-level interpretations of semantic meaning. This rich annotation structure enables systematic evaluation of both human and machine understanding of combinational creativity at multiple cognitive levels.

\subsection{Task Setups}

We design three tasks corresponding to the levels of our \ac{iei} framework, each probing progressively deeper aspects of combinational creativity understanding:

\paragraph{Identification}

This task probes the basic object recognitions in a visual mashup image. These basic objects serve as the source components that can be combined to create the final image. Given a visual mashup image, models must identify the two primary objects that have been combined, requiring both visual perception and conceptual understanding. We assess performance using precision and recall. Our system uses GPT-4o to perform object matching by incorporating the homogeneous meaning sets that were created during annotation, along with appropriate instructions/examples to guide the process.

\paragraph{Explanation}

This task examines the understanding of the combinatorial mechanisms underlying visual mashups. Models must identify specific attributes that allow for feasible combinations in a visual mashup image. Typically, these are attributes shared by the two basic objects, such as similar shapes, functions, or semantic properties that enable meaningful integration. When evaluating performance in recognizing the attributes, correctness is assessed by comparing the selected attributes with a predefined list of ground-truth attributes for each image. We use precision to measure this performance.

\paragraph{Implication}

This task assesses deeper semantic understanding by requiring models to interpret what the combination in a visual mashup image might be intended to convey and express. This includes identifying underlying themes, emotions, cultural references, or conceptual messages embedded in the creative combination. The task demands not just recognition of combined elements, but comprehension of emergent meaning that arises from their integration. To assess both models' and humans' capabilities in this task, we perform pairwise comparisons among the multiple implications associated with each image. To automate the evaluation pipeline, we use GPT-4o as an adjudicator. We validate the reliability of this method by having two human experts annotate preferences for 200 comparison results, achieving consistency rates of 85.1\% and 81.6\% with GPT-4o. This automated evaluation technique for open-ended responses has demonstrated robustness in previous research \citep{ji2024pku,chiang2024chatbot}.

\begin{table}[t!]
    \centering
    \small
    \caption{\textbf{Model performance on comprehension tasks.} Models are evaluated on identification (\textbf{P}/\textbf{R}), explanation (\textbf{P}), and implication \textbf{WR} metrics for combinational creativity understanding. \textbf{P} refers to precision, \textbf{R} to recall, and \textbf{WR} to winning rate. Human performance (gray row) serves as the baseline for \acsp{vlm}.}
    \setlength{\tabcolsep}{3pt}
    \resizebox{\linewidth}{!}{%
        \begin{tabular}{lcccc}
            \toprule
            \multirow{2}{*}{Models}                           & \multicolumn{2}{c}{Identification}                          & Explanation           & Implication \\   
            \cmidrule{2-3}                                    & \textbf{P}$\uparrow$                & \textbf{R}$\uparrow$  & \textbf{P}$\uparrow$  & \textbf{WR}$\uparrow$ \\ 
            \midrule
            \rowcolor{gray!20} Human Experts                  & -                                   & -                     & -                     & \textbf{78.3} \\
            \rowcolor{gray!20} Average People                 & 53.42                               & 70.33                 & 69.89                 & 51.0 \\
            \midrule
            GPT-4o \citep{openai2024gpt4o}                    & \textbf{75.67}                      & \textbf{85.00}        & \textbf{74.19}        & 73.5 \\
            GPT-4V \citep{openai2023gpt4V}                    & 60.83                               & 75.00                 & 63.44                 & 71.9 \\
            Gemini-1.5-Pro \citep{team2024gemini}             & 73.67                               & 81.33                 & 54.34                 & 71.7 \\
            Claude-3.5-Sonnet \citep{anthropic2024claude3.5}  & 60.08                               & 74.83                 & \textbf{74.19}        & 62.9 \\
            Claude-3-Opus \citep{anthropic2024claude3}        & 63.17                               & 72.50                 & 65.59                 & 39.2 \\
            LLaVA-1.6-34B \citep{liu2024improved}             & 64.67                               & 72.17                 & 62.37                 & 40.6 \\
            LLaVA-1.6-13B \citep{liu2024improved}             & 60.33                               & 67.33                 & 40.86                 & 34.3 \\
            LLaVA-1.6-7B \citep{liu2024improved}              & 50.33                               & 57.83                 & 48.39                 & 20.8 \\
            LLaVA-1.5-7B \citep{liu2024visual}                & 49.62                               & 63.00                 & 43.01                 & 20.1 \\
            MiniCPM \citep{hu2024minicpm}                     & 64.40                               & 72.33                 & 50.54                 & 41.7 \\
            Qwen-VL-Chat \citep{bai2023qwen}                  & 55.50                               & 62.50                 & 65.59                 & 41.9 \\ 
            \bottomrule
        \end{tabular}%
    }%
    \label{tab:model_comprehension}
\end{table}

\subsection{Experimental Setups}

We establish human baselines and evaluate a comprehensive set of models to assess combinational creativity understanding.

\paragraph{Human Baseline}

To establish performance benchmarks for average human understanding of combinational creativity, we recruited participants through the Prolific platform, the protocol of which was approved by a university IRB. These participants followed identical annotation guidelines to those used in creating our expert-annotated dataset, ensuring direct comparability between expert and average human performance.

\paragraph{Models}

We evaluate a diverse array of 11 models \citep{openai2024gpt4o,openai2023gpt4V,anthropic2024claude3,anthropic2024claude3.5,liu2024visual,liu2024improved,hu2024minicpm,bai2023qwen} spanning different capabilities and architectures. Our selection includes closed-source models like GPT-4o, Gemini-1.5-Pro, and Claude-3.5-Sonnet, as well as open-source alternatives such as LLaVA \citep{liu2024visual,liu2024improved} and Qwen \citep{bai2023qwen}. The models cover various parameter sizes (34B, 13B, 8B, and 7B) and different \ac{llm} architectures (Yi-34B, Llama3-8B, Vicuna-7B, and Qwen-7B), enabling comprehensive evaluation across model scales, architectures, and accessibility levels.

\subsection{Results}

\paragraph{Identification}

As shown in \cref{tab:model_comprehension}, GPT-4o demonstrates superior performance in identifying the basic objects in visual mashups, achieving the highest precision and recall scores. Gemini-1.5-Pro follows as a strong performer with 73.67\% precision and 81.33\% recall. Average human participants show moderate competency in this task, achieving 53.42\% precision and 70.33\% recall, indicating that even basic object identification in creative combinations presents notable challenges.

\paragraph{Explanation}

In analyzing combinatorial mechanisms, GPT-4o maintains its leading position, while Claude-3.5-Sonnet shows particularly interesting results---despite moderate performance in identification, it matches GPT-4o's top performance in explanation. A notable observation is Gemini-1.5-Pro's performance disparity, showing strong identification capabilities but relatively weaker explanation abilities. Average humans demonstrate strong explanatory capabilities, achieving 69.89\% precision and ranking third, suggesting humans' natural ability to understand combinatorial mechanisms.

\paragraph{Implication}

\begin{figure}[t!]
    \centering
    \includegraphics[width=\linewidth]{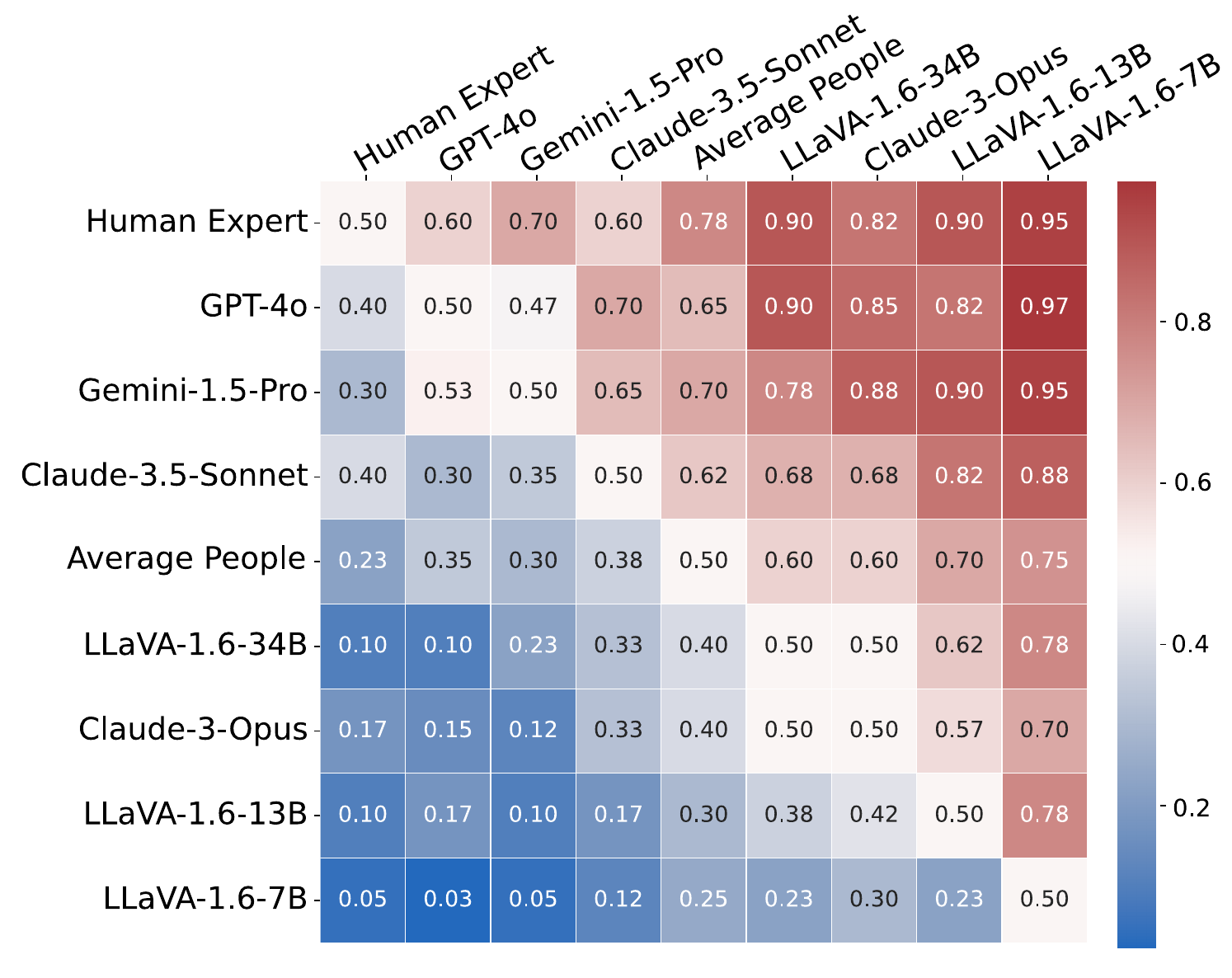}
    \caption{\textbf{Pairwise model comparison on implication task.} The heatmap displays winning probabilities, where each cell $(i,j)$ shows the win rate of row model $i$ \vs column model $j$. Darker red indicates higher win rates, while darker blue represents lower win rates.}
    \label{fig:intermap} 
\end{figure}

The implication task reveals the most distinctive performance patterns. Human experts demonstrate superior understanding with a 78.3\% winning rate, significantly outperforming all models. Average participants achieve a 51\% winning rate, positioning in the middle range of performance. Among models, GPT-4o leads the artificial systems, followed by GPT-4V and Gemini-1.5-Pro, all achieving winning rates above 70\%. These patterns are further confirmed through pair-wise comparisons (\cref{fig:intermap}), where human experts consistently maintain the highest winning rate, while average participants typically fall within the middle performance range across model comparisons.

\subsection{Analysis and Discussions}

Based on our comprehensive analysis of \cref{tab:model_comprehension,fig:intermap} and experimental data, we present three key findings.

\paragraph{State-of-the-art \acp{vlm} have achieved above-average human performance in comprehension tasks}

Our results demonstrate that leading models like GPT-4o and Claude-3.5-Sonnet have surpassed average human performance across all three tasks in understanding combinational creativity.

\paragraph{Fusion combinational type shows higher cognition rates than replacement}

\begin{figure}[t!]
    \centering
    \begin{subfigure}[b]{0.49\linewidth}
        \centering
        \includegraphics[width=\linewidth]{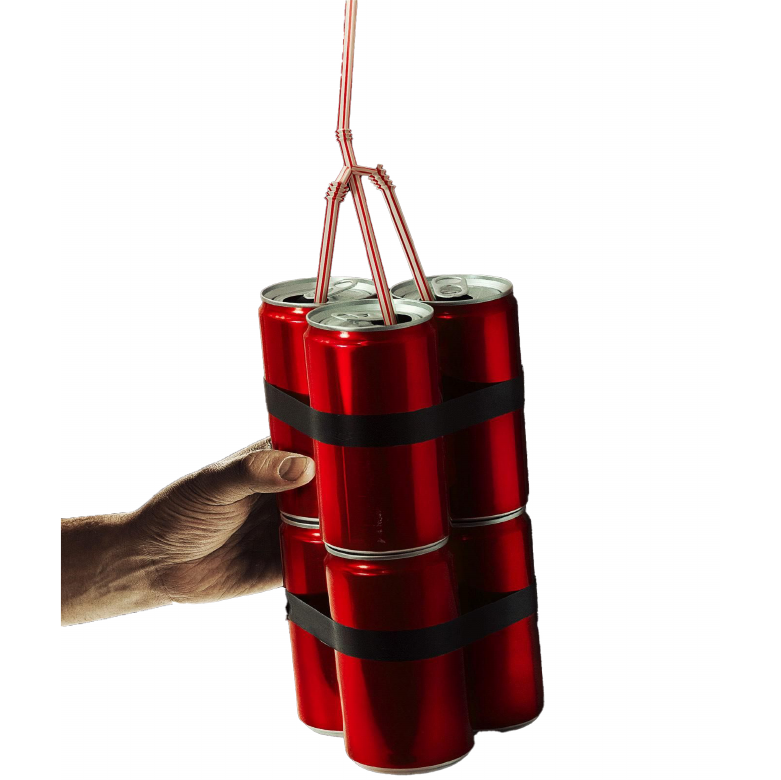}
        \caption{The \textbf{replacement} of dynamite with soda cans.}
        \label{fig:apfaila}
    \end{subfigure}%
    \hfill
    \begin{subfigure}[b]{0.49\linewidth}
        \centering
        \includegraphics[width=\linewidth]{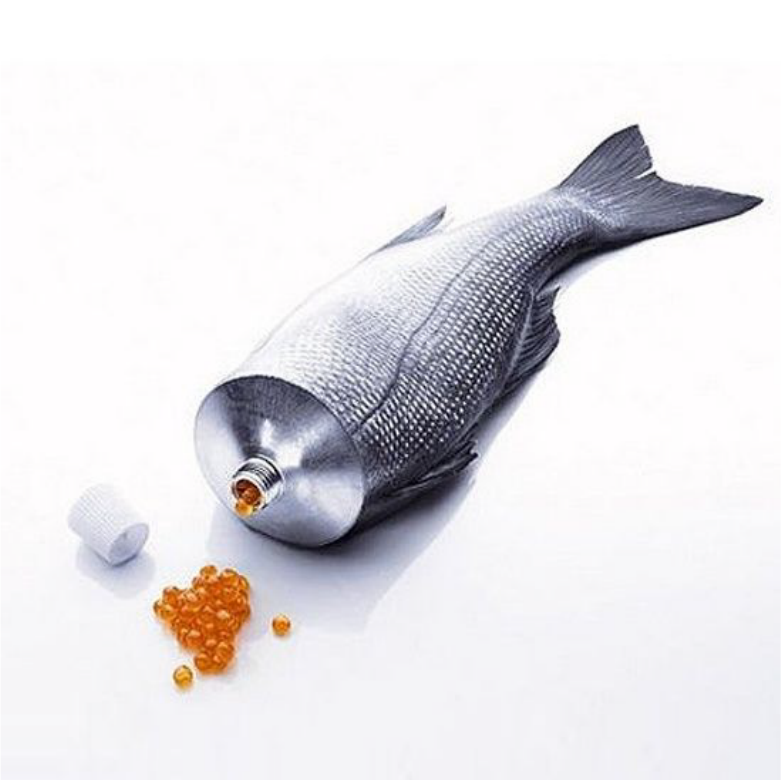}
        \caption{The \textbf{fusion} between a fish and a toothpaste.}
        \label{fig:apfailb}
    \end{subfigure}%
    \label{fig:apfail}
    \caption{\textbf{Two types of combination in comprehension tasks.} (a) \textbf{Replacement} maintains functional or visual similarity while substituting for safer or more accessible alternatives. (b) \textbf{Fusion} merges two unrelated concepts to create a novel composite that inherits properties from both sources.}
    \vspace{-1mm}
\end{figure}

\begin{figure}[t!]
    \centering
    \includegraphics[width=\linewidth]{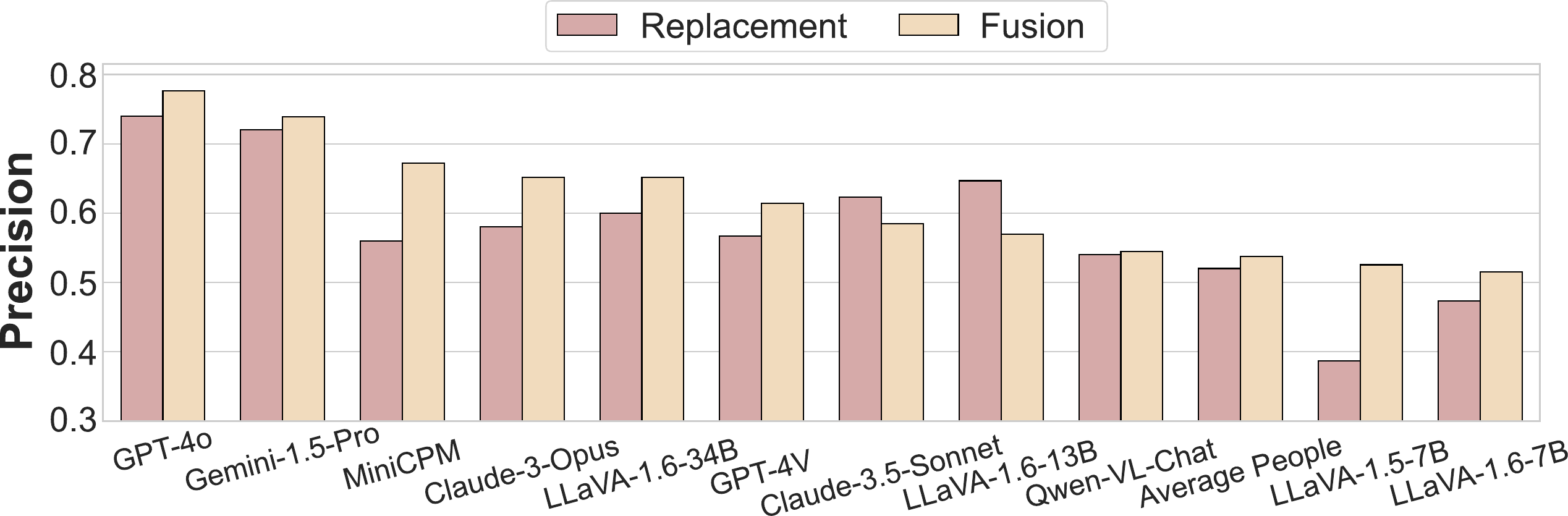}
    \caption{\textbf{Precision in identification task by combination type: replacement \vs fusion.} Precision metrics across combination categories show consistently higher precision for replacement-based combinations compared to fusion-based ones.}
    \label{fig:repfus}
    \vspace{-1mm}
\end{figure}

Our visual mashups exhibit two distinct combination types: fusion and replacement. Fusion combinations merge characteristics from different basic objects, creating a blended object that preserves visual traits from both sources (see \cref{fig:apfailb}). Replacement combinations involve one basic object substituting another within the structure (see \cref{fig:apfaila}). Analysis of 11 models reveals higher precision metrics for fusion combinations in 9 models (see \cref{fig:repfus}), mirroring typical human performance patterns. This disparity stems from the inherent nature of these combinations. Fusion combinations offer more readily identifiable characteristics from both basic objects, facilitating easier deconstruction. Conversely, replacement combinations demand a deeper understanding of contextual relationships or shared characteristics, as they predominantly display traits from one basic object while retaining only minimal characteristics from the other.

\paragraph{Semantic attribute range and depth define implication quality}

Our comparison of expert and average human implications reveals two critical factors determining interpretation quality: semantic attribute range and depth. For example, in \cref{fig:apfaila}, an average person's simple interpretation ``A taste explosion'' contrasts with the expert's more nuanced analysis: ``Drinking soda in excess is like detonating explosives within your body, slowly destroying your health.'' The expert interpretation incorporates additional semantic attributes like ``in excess'' and ``destroy health,'' extending beyond the basic ``taste'' attribute. Similarly, in \cref{fig:apfailb}, the profound metaphorical interpretation ``Humanity relentlessly extracts resources'' transcends the more literal observation ``The fish is killed for its roe and packaged for human consumption.'' While the average interpretation correctly identifies ``human consumption,'' it doesn't explore deeper semantic connections between fish consumption and resource extraction. These examples illustrate how varying ranges and depths of semantic attributes lead to distinctly different levels of interpretation.

\section{Experiment 2: How to Induce \texorpdfstring{\acp{vlm}}{}' Combinational Creativity?}

Building on conceptual blending theory \citep{fauconnier2003conceptual}, we hypothesize that explicitly incorporating a three-step thinking process would induce or enhance \acp{vlm}' ability to generate products exhibiting combinational creativity.

\subsection{Task Setups}

This task requires \acp{vlm} to generate creative images by combining provided objects to convey specific themes (see \cref{fig:task}\textcolor{red}{b}). These thematic constraints are essential, as they enable systematic comparison of both novelty and usefulness in the generated images, two key dimensions in evaluating creative products.

Drawing from our annotated visual mashups dataset, we structured the input information across three levels: \one Identification: the input objects available for potential combination; \two Explanation: the shared attributes that may guide the combination process; \three Implication: the semantic meaning of the creative output, which serves as a source for the theme. 

Our central research question examines whether explicitly incorporating a three-step \ac{iei} thinking process would enhance the combinational creativity manifested in the generated images. To investigate this, we compare three conditions:
\begin{enumerate}[leftmargin=*,noitemsep,nolistsep]
    \item \textbf{Identification+Implication (\baseline{})}: Models are instructed to creatively combine the input objects while considering the theme, using chain-of-thought prompting \citep{wei2022chain} without any external guidance in the combinational process.
    \item \textbf{Identification+Explanation+Implication (\cc{})}: Models are explicitly instructed to follow a three-step \ac{iei} thinking process, with additional guidance focusing on the potential shared attributes.
    \item \textbf{Human Expert}: The original artist-generated visual mashups that serve as our reference point.
\end{enumerate}

For both \baseline{} and \cc{} conditions, we begin by providing the inputs as prompts to GPT-4o. This process involves combining basic objects at the conceptual level and generating appropriate prompts for text-to-image generative models. These generated prompts are then used to drive the image-generation process.

\subsection{Experimental Setups}

\paragraph{Human Evaluations}

To assess the effectiveness of our approach, we recruited 50 participants through Prolific for preference evaluation. Participants were tasked with ranking image triplets (i.e., \textbf{Human Experts}, \baseline{}, and \cc{}) from different experimental conditions based on two key criteria: image novelty and effectiveness of conveying the theme. To minimize potential biases, we randomized the presentation order of triplets across participants.

\paragraph{Models}

We evaluated our method across five state-of-the-art text-to-image generation models, encompassing both closed- and open-source models. The selected models included Midjourney \citep{midjourney}, Flux-1.1-pro \citep{black2024fluxp}, DALLE-3 \citep{dalle3}, Flux-1.0-dev \citep{black2024flux}, and Stable-Diffusion-3-medium \citep{esser2024scaling}. This diverse selection of models affords an examination of the generalizability of our approach across different generative models.

\subsection{Results}

\begin{figure}[t!]
    \centering
    \includegraphics[width=\linewidth]{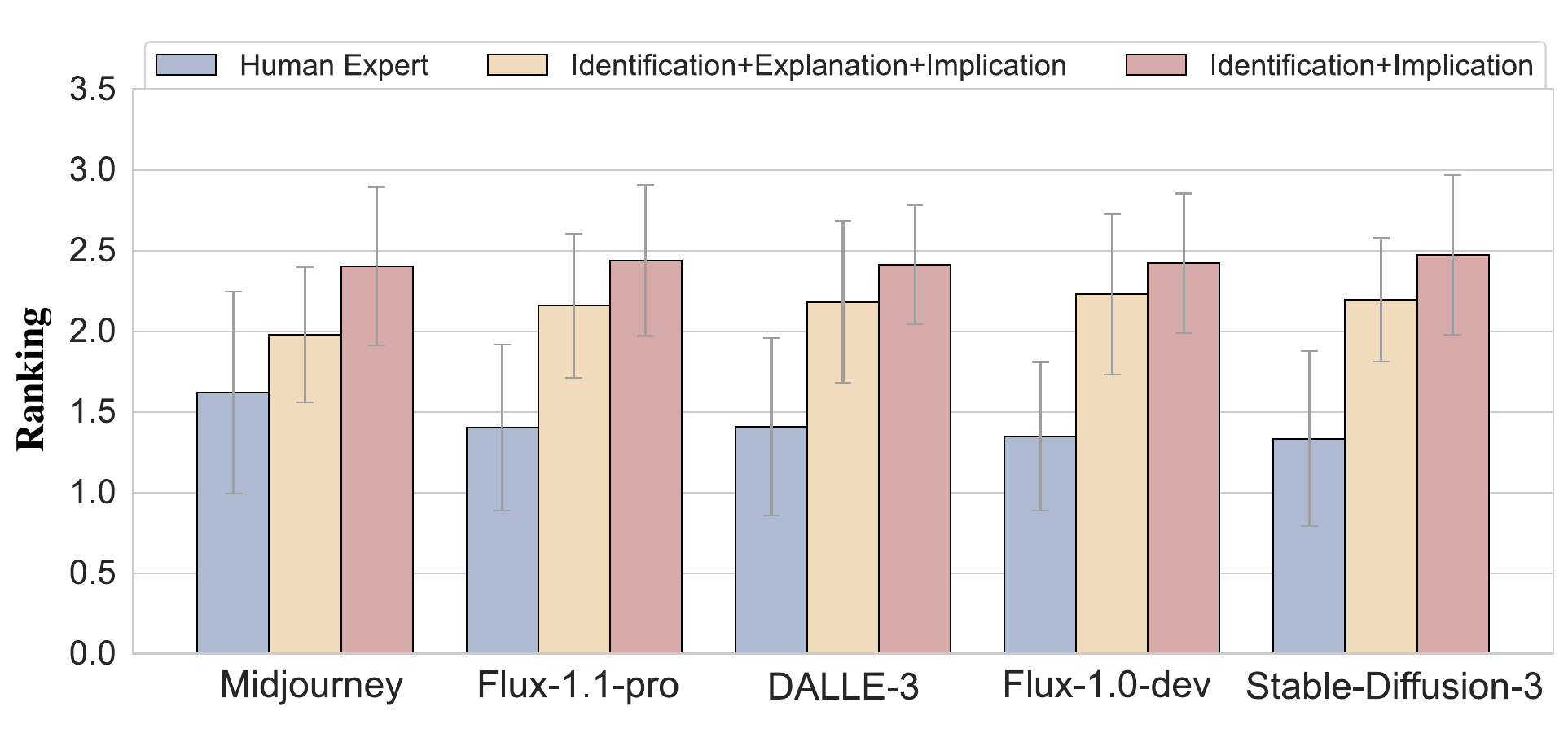}
    \caption{\textbf{Human preference rankings across in generation task.} We compare rankings from Human Expert, \cc{}, and \baseline{} assessments. (Lower scores indicate better performance; error bars show variance.)}
    \label{fig:genrank} 
    \vspace{-1mm}
\end{figure}

\cref{fig:genrank} presents human evaluation results of image triplets across different generation conditions. The Friedman test revealed significant differences in ranking scores among the three conditions for all models ($p < 0.0001$). Post-hoc analysis revealed that \cc{}-generated images significantly outperformed \baseline{}-generated images, while artist-generated images maintained superior performance overall.

Among the evaluated models, Midjourney demonstrated the strongest performance. The \cc{} and \baseline{} approaches achieved mean rankings of $1.98$ and $2.40$, respectively, leading ahead of other models' performance. Notably, human-expert artworks showed the highest ranking variance compared to Midjourney-generated images, suggesting more diverse opinions on human-created art. For less sophisticated models, participants more consistently identified the superiority of human-created images, indicating that these models produce outputs more readily distinguishable from human artworks.

\cref{fig:winrate} illustrates the winning rates comparing \baseline{} and \cc{} conditions across models. Midjourney exhibited superior performance in both conditions, with \cc{} enhancing its winning rate from $0.26$ to $0.35$. All models demonstrated improved winning rates under the \cc{} condition compared to their \baseline{} performance, though improvement magnitudes varied. The performance differential between conditions diminished for lower-performing models, with Stable-Diffusion-3 showing minimal improvement from \cc{} enhancement.

\vspace{-1mm}
\subsection{Analysis and Discussions}

\paragraph{\acs{iei} enhances creativity independent of prompt length}

\begin{figure}[t!]
    \centering
    \includegraphics[width=0.91\linewidth]{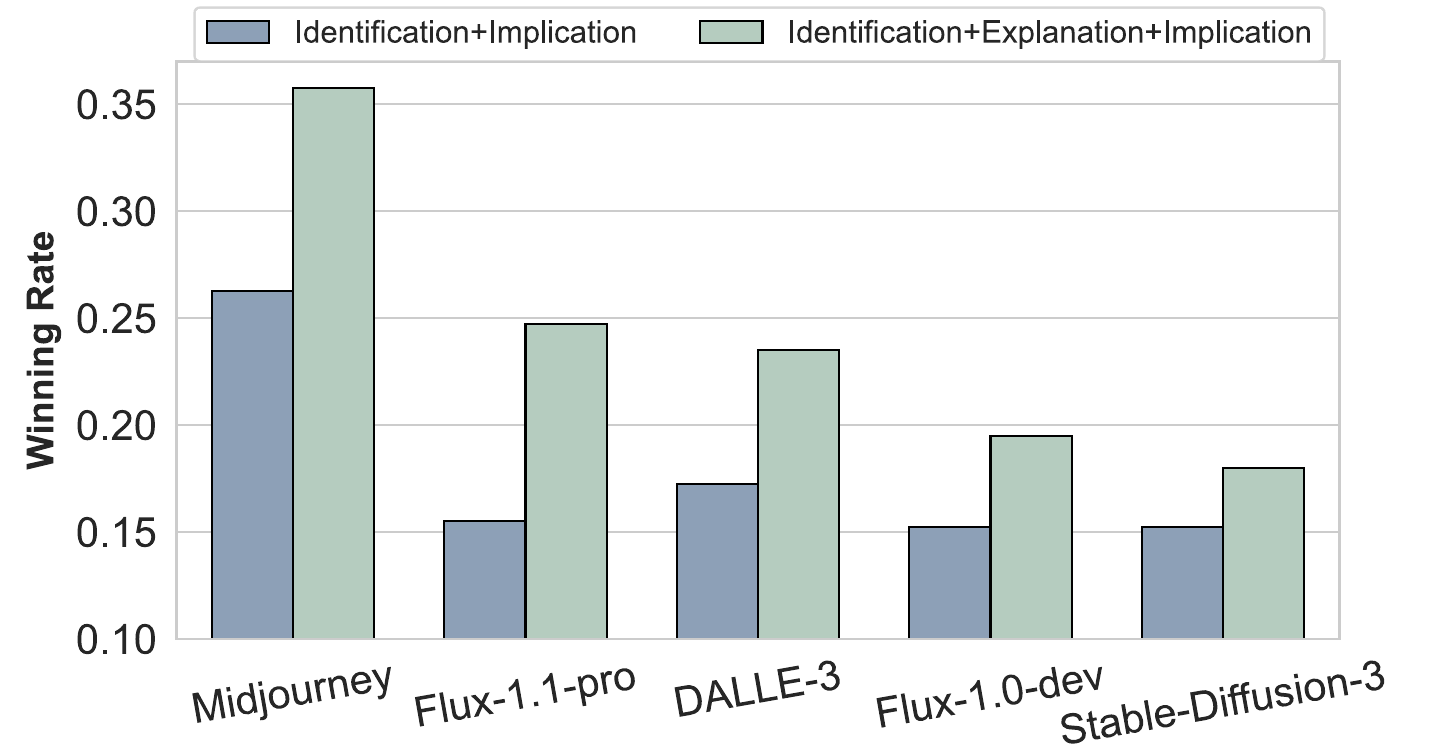}
    \caption{\textbf{Winning rates against human experts in generation task.} Win percentages comparing \baseline{} and \cc{} settings against human expert assessment, demonstrating relative performance differences between the two approaches in human evaluation.}
    \label{fig:winrate}
    \vspace{-1mm}
\end{figure}

The primary distinction between \cc{} and \baseline{} conditions lies in incorporating the explanation level of combinational creativity as a conceptual guiding mechanism. As evidenced in \cref{fig:genrank,fig:winrate}, \cc{} yielded significantly superior results compared to \baseline{}. This improvement was particularly pronounced in more sophisticated models, which demonstrated greater benefits from the additional guidance. We attribute this pattern to these models' enhanced capability to process and leverage the increased conceptual complexity in generating higher-quality images.

To verify that these improvements were not simply due to more verbose instructions, we conducted analyses of prompt lengths. An independent samples t-test comparing prompt lengths revealed no significant difference between \baseline{} (M = 487.95) and \cc{} (M = 514.53) conditions ($t=-0.84, p=0.404$). To further investigate this relationship, we examined all instances where \cc{} outperformed \baseline{}. In 46.0\% of these successful cases, \cc{} prompts were shorter than their \baseline{} counterparts, demonstrating that the explanation level's effectiveness stems not from increased verbosity.

These findings suggest that introducing the three-step \ac{iei} thinking process, especially the explanation level of combinational creativity, provides essential conceptual guidance that facilitates more meaningful concept integration. The improvement in creative output appears to derive from the structural and semantic guidance provided by the explanation level rather than from simply providing more detailed instructions to the image generation models.

\paragraph{Text-to-image models are the bottleneck in generating visual combinational creativity images}

Our analysis reveals a significant disparity between conceptual combination and visual execution capabilities in current AI systems. Using GPT-4o for concept-level object combination showed promising results, with 90\% (36 out of 40) of DALLE-3 generated examples successfully integrating basic objects and effectively conveying intended themes. However, the translation from conceptual description to visual output proved more challenging, with 27.8\% (10 out of 36) of well-crafted descriptions failing to produce satisfactory visual results.

This performance gap illuminates a crucial challenge. While \ac{llm} demonstrate strong conceptual combination abilities, text-to-image models often struggle to execute these creative concepts faithfully. This observation aligns with our experimental findings, in which more sophisticated models showed greater improvement with enhanced instructions, suggesting that the ability to follow instructions significantly impacts the quality of visual generation. While these results highlight the potential of \ac{llm} as creative design tools for skilled human artists, they also indicate that advancing text-to-image models' instruction-following capabilities remains a critical priority for achieving fully automated creative processes.

\vspace{-1mm}
\section{Conclusion}

This work advances the understanding of combinational creativity in \acp{vlm} through a three-level \ac{iei} framework. We contribute a framework-annotated dataset of visual mashups for probing understanding capabilities, and an \ac{iei}-based process that enhances generative performance in \acp{vlm}. Experiments reveal that while state-of-the-art models can surpass average human performance in recognizing combinational creativity, they still fall short of the expert level. We show that incorporating the combinational process into model operation enhances creative output, validating our framework's utility. This work establishes a foundation for analyzing combinational creativity, with our dataset and evaluation methodologies serving as resources for future research in creative \ac{ai} systems.

\section{Acknowledgments}

We gratefully acknowledge Chengdong Ma and Qinghao Wang for their valuable assistance with data exploration, Yujia Peng for her suggestions on the human study, and Zhen Chen for her efforts in figure preparation, and Hongjie Li for his advice on project website. This work is supported in part by the National Science and Technology Major Project (2022ZD0114900), the National Natural Science Foundation of China (62376031), the Beijing Nova Program, the State Key Lab of General AI at Peking University, the PKU-BingJi Joint Laboratory for Artificial Intelligence, and the National Comprehensive Experimental Base for Governance of Intelligent Society, Wuhan East Lake High-Tech Development Zone.

\bibliographystyle{apacite}
\balance
\renewcommand\bibliographytypesize{\small}
\setlength{\bibleftmargin}{.125in}
\setlength{\bibindent}{-\bibleftmargin}
\bibliography{reference_header,reference}

\end{document}